\documentclass{sigchi}
\usepackage{graphicx}
\usepackage{url}
\usepackage[table]{xcolor}    

\usepackage{xcolor,cancel}

\def\pprw{8.5in}
\def\pprh{11in}

\begin{document}

\title{Delving Deeper into MOOC Student Dropout Prediction}
\numberofauthors{3} 
%
\author{
  \alignauthor{Jacob Whitehill\\
    \affaddr{Worcester Polytechnic Institute}\\
    \affaddr{Worcester, MA, USA}\\
    \email{jrwhitehill@wpi.edu}}\\
  \alignauthor{Kiran Mohan\\
    \affaddr{Worcester Polytechnic Institute}\\
    \affaddr{Worcester, MA, USA}\\
    \email{kmohan@wpi.edu}}\\
  \alignauthor{Daniel Seaton\\
    \affaddr{Harvard University}\\
    \affaddr{Cambridge, MA, USA}\\
    \email{daniel\_seaton@harvard.edu}}\\
  \alignauthor{Yigal Rosen\\
    \affaddr{Harvard University}\\
    \affaddr{Cambridge, MA, USA}\\
    \email{yigal\_rosen@harvard.edu}}\\
  \alignauthor{Dustin Tingley\\
    \affaddr{Harvard University}\\
    \affaddr{Cambridge, MA, USA}\\
    \email{dtingley@gov.harvard.edu}}\\
%
%
}
\date{26 September 2016}

\maketitle
\begin{abstract}
In order to obtain reliable accuracy estimates 
for automatic MOOC dropout predictors, it is important to train and test them in a manner consistent
with how they will be used in practice. Yet most prior research  on MOOC dropout prediction
has measured test accuracy on the same course used for training the classifier, which
can lead to overly optimistic accuracy estimates.
In order to understand better how accuracy is affected by the training+testing regime,
we compared the accuracy of a standard dropout prediction architecture (clickstream features + logistic regression)
across 4 different training paradigms.
Results suggest that  (1) training and testing on the same course (``post-hoc'')
can overestimate accuracy by several percentage points;
(2) dropout classifiers trained on proxy labels based on students' \emph{persistence}
are surprisingly competitive with post-hoc training ($87.33\%$ versus $90.20\%$ AUC
averaged over 8 weeks of 40 HarvardX MOOCs); and (3)
classifier performance does not vary significantly with the academic discipline. Finally, we 
also research new dropout prediction architectures based on deep, fully-connected, feed-forward
neural networks and find that (4) networks with as many as 5 hidden layers can statistically
significantly increase test accuracy over that of logistic regression.
\end{abstract}

\section{Introduction and related work}
As the number and diversity of massive open online courses (MOOCs) continues to grow, researchers, teachers, and educational
technologists are starting to explore innovative ways of helping more students who participate in these courses to persist
longer and learn more. While some interventions (e.g., email reminders \cite{WhitehillEtAl2015}) may incur very little cost
once the core infrastructure for conducting them is in place, other types of interventions -- such as a course
staff-member actively reaching out to a learner
to ask her/him how she/he is doing \cite{simpson2004impact,firmin2014case},
or an instructor responding substantively within a discussion forum to a comment the student
had written \cite{WongEtAl2016} -- may have high variable costs and be very time-intensive. An automatic detector that could predict automatically
-- based on demographics or by analyzing subtle cues in a learner's interaction history with the MOOC courseware -- 
which learners are in danger of ``dropping out'' from the MOOC and which will likely succeed, would be a valuable tool for making
smart decisions about who receives a particular intervention.

{\bf Related work}: Within the fields of learning analytics and educational data mining,
the possibility of creating automatic MOOC dropout detectors has generated considerable interest within the past few years.
Existing approaches vary across several dimensions including the \emph{features}
used for classification, the \emph{architecture} used for training and testing, and the \emph{training set} used to optimize
the classifier parameters -- see Table \ref{tbl:lit_review} for a synopsis of prior work.
The most popular feature representations include clickstream logs, natural language processing (NLP)
of discussion forum content, and social network metrics. In terms of architecture, most prior approaches have
used generalized linear models (including logistic regression and linear SVMs), survival analysis (e.g., Cox proportional hazard
model), and logistic regression. A third dimension of variability is the \emph{training setting} that
describes the \emph{source} of the training data -- e.g., the same course, a prior instance of the same course,
or a different course altogether -- relative to the \emph{target use} of the classifier once it has been
trained: most research on MOOC dropout prediction to-date has focused
on training and testing on data sampled from the same MOOC.
\begin{table*}
\begin{center}
\begin{tabular}{l|c|l|l|l}
\multicolumn{5}{c}{\bf Survey of Prior Research on MOOC Dropout Prediction}\\\hline
{\bf Study} & {\bf \#MOOCs} & {\bf Features} & {\bf Architecture} & {\bf Training setting} \\\hline\hline
Balakrishnan \& Coetzee~\cite{BalakrishnanBerkeley2013} & 1 & Clickstream & HMM + SVM & Same course\\\hline
Boyer \& Veeramachaneni~\cite{boyer2015transfer} & 3 & Clickstream & TL+LR & \begin{tabular}{@{}l}Different offering\\In-situ\end{tabular} \\\hline
Coleman et al.~\cite{ColemanEtAl2015} & 1 & Clickstream & LDA+LR & Same course\\\hline
Crossley et al.~\cite{crossley2016combining} & 1 & Clickstream; NLP & DFA & Same course\\\hline
Fei \& Yeung~\cite{fei2015temporal} & 2 & Clickstream & RNN & Same course\\\hline
He et al.~\cite{HeEtAl2015} & 2 & Clickstream & Smoothed LR  & Different offering\\\hline
Jiang et al.~\cite{JiangEtAl2014} & 1 & Social network; grades & LR & Same course\\\hline
Kizilcec et al.~\cite{KizilcecHalawa2015,HalawaEtAl2014} & 20 & Clickstream & LR & \begin{tabular}{@{}l}Different course\\Same course\end{tabular} \\\hline
Kloft et al.~\cite{KloftEtAl2014} & 1 & Clickstream & SVM & Same course \\\hline
Koedinger et al.~\cite{koedinger2015learning} & 1 & Clickstream; grades & LR & Same course \\ \hline
Robinson et al.~\cite{robinson2016forecasting} & 1 & Survey; NLP & LR & Same course\\\hline
Rose et al.~\cite{YangEtAl2014,rose2014social} & 1 & Forum; social network & SA & Same course \\ \hline
Stein \& Allione~\cite{SteinAllione2014} & 1 & Clickstream; survey & SA & Same course \\ \hline
Taylor et al.~\cite{TaylorEtAl2014} & 1 & Clickstream & LR & Same course \\\hline
Whitehill et al.~\cite{WhitehillEtAl2015} & 10 & Clickstream & LR & Different course \\\hline
Xing et al.~\cite{xing2016temporal} & 1 & Clickstream; social network & PCA+\{BN,DT\} & Same course\\\hline
Ye \& Biswas~\cite{YeBiswas2014} & 1 & Clickstream & LR & Same course \\ \hline
{\bf Our paper} & 40 & Clickstream & \{LR, DNN\} & \begin{tabular}{@{}l}Same course\\In-situ\\Different course\end{tabular} \\\hline
\end{tabular}
\end{center}
\caption{Survey of prior literature on MOOC dropout prediction. For the architecture, we use the following
abbreviations: Bayesian network (BN), decision tree (DT), deep neural network (DNN), discriminant function
analysis (DFA), hidden Markov model (HMM), latent Dirichlet allocation (LDA), logistic regression (LR),
principal component analysis (PCA), recurrent neural network (RNN), support vector machine (SVM), survival analysis (SA),
and transfer learning (TL). Architecture $a+b$ means methods $a$ and $b$ were used in conjunction; $\{a,b\}$ means
that $a$ or $b$  were used as alternatives.}
\label{tbl:lit_review}
\end{table*}

When designing and implementing educational interventions that depend on automatic MOOC dropout predictors to
predict which students are in danger of failing, it is important  to know how accurate such predictors are.
In order to obtain reliable accuracy estimates, it is crucial to ensure that the method used for evaluating
the classifier is consistent with the manner it will be used for the actual intervention.
Live MOOC interventions require dropout predictors that are operational at or near the \emph{beginning} of a MOOC
while students can still benefit from receiving an intervention. But producing such a predictor
can be difficult because the target values which
indicate whether each student dropped out or completed the MOOC, and which are usually required for training a classifier
with standard supervised learning approaches, typically become available only at the \emph{end} of a MOOC --
at which point any intervention
is moot.

To-date, most prior research on MOOC dropout detection has largely ignored this issue and
both trained and tested on student data sampled from the same course (``post-hoc''), likely because dropout prediction
was a new field and the focus was on exploring different feature representations and classification architectures.
The study we present in this paper seeks to fill this gap.

{\bf Contributions}: (1) In this paper
we compare the accuracy of a standard dropout prediction architecture -- 
clickstream features classified by logistic regression -- across a variety of different training settings
in order to understand better the tradeoff between accuracy and practical deployability of the classifier.
The dataset we use for training and evaluation consists of 40 popular MOOCs from HarvardX that span
a variety of academic disciplines including social science, humanities, STEM, and health sciences.
As part of this evaluation, (2) we also explore a novel scheme (similar to the \emph{in-situ} approach proposed by
\cite{boyer2015transfer}) for training a classifier \emph{while} a course is ongoing using \emph{proxy labels}
and show that its performance is surprisingly good even compared to the overly optimistic \emph{post-hoc} training
paradigm. Finally, (3) we propose and evaluate a novel MOOC dropout architecture based on fully-connected, feed-forward neural
networks that are trained at each week in a MOOC. We provide evidence that much deeper (5 hidden layers) feature
representations than have previously been explored can lead to statistically significantly higher prediction accuracy.


\section{Dataset}
The experiments and analyses in this paper are based on data from 40 MOOCs from HarvardX -- see Table \ref{tbl:mooc_stats} which
lists the code name of each course; the year and term (T1, T2, and T3 are  northern hemisphere spring, summer, and fall, respectively)
when the course was offered; the academic discipline; number of registered participants;  and the number of participants who certified.
Each MOOC has a \emph{launch date} (which we call $T_{0\%}$) when the first materials -- e.g., lecture videos, discussion forums, etc. -- are released,
as well as an \emph{end date} when certificates are issued.

\begin{table}
\footnotesize
\begin{tabular}{l|l|c|r|r}
\multicolumn{5}{c}{\bf List of MOOCs}\\\hline
{\bf Course} & {\bf Year} & {\bf Field} & {\bf \#Part.} & {\bf \#Cert.}\\\hline
1368.1x & 2014T3 & SocialSci  & 2335 & 387\\
1368.3x & 2015T2 & SocialSci  & 1278 & 395\\
AmPoX.4 & 2015T1 & Hum  & 4348 & 344\\
AmPoX.5 & 2015T2 & Hum  & 2296 & 195\\
AmPoX.6 & 2016T2 & SocialSci  & 2810 & 45\\
BUS5.1x & 2014T1 & SocialSci  & 9386 & 575\\
BUS5.1x & 2015T3 & SocialSci  & 4363 & 173\\
EMC2x & 2015T1 & SocialSci  & 8257 & 196\\
ENGSCI137x & 2016T2 & STEM  & 7356 & 339\\
ER22.1x & 2014T1 & SocialSci  & 23972 & 2375\\
GSE2x & 2014T2 & SocialSci  & 23740 & 3847\\
HDS3221.1x & 2016T1 & Hum  & 17949 & 258\\
HKS101A & 2015T3 & SocialSci  & 9849 & 599\\
HKS211.2x & 2015T3 & SocialSci  & 3872 & 791\\
HLS2x & 2015T1 & SocialSci  & 10814 & 2190\\
HLS3x & 2015T1 & SocialSci  & 2468 & 354\\
MCB63X & 2015T3 & STEM  & 11225 & 561\\
MCB64.1x & 2016T2 & STEM  & 6984 & 268\\
PH210x & 2014T1 & HealthSci  & 6390 & 745\\
PH231x & 2016T1 & HealthSci  & 4372 & 414\\
PH525.1x & 2015T1 & HealthSci  & 14300 & 1804\\
PH525.2x & 2015T1 & HealthSci  & 5420 & 869\\
PH525.3x & 2015T1 & HealthSci  & 4690 & 464\\
PH525.5x & 2015T3 & HealthSci  & 1410 & 120\\
PH525x & 2014T1 & HealthSci  & 10705 & 628\\
PH556 & 2015T3 & HealthSci  & 6346 & 900\\
SPU30x & 2014T2 & STEM  & 5611 & 941\\
SW12.10x & 2015T1 & SocialSci  & 5176 & 1246\\
SW12.2x & 2014T1 & SocialSci  & 8660 & 2085\\
SW12.3x & 2014T1 & SocialSci  & 5051 & 1825\\
SW12.4x & 2014T1 & SocialSci  & 4979 & 1527\\
SW12.5x & 2014T2 & SocialSci  & 3466 & 1458\\
SW12.6x & 2014T2 & SocialSci  & 4011 & 1383\\
SW12.7x & 2014T3 & SocialSci  & 3749 & 1391\\
SW12.8x & 2014T3 & SocialSci  & 3677 & 1392\\
SW12.9x & 2014T3 & SocialSci  & 3569 & 1380\\
SW25x & 2014T1 & HealthSci  & 6444 & 1246\\
SW25x.1 & 2015T1 & HealthSci  & 3377 & 453\\
USW30x & 2016T1 & SocialSci  & 3135 & 350\\
USW30x & 2014T2 & SocialSci  & 8245 & 1045
\normalsize
\end{tabular}
\caption{
List of the $40$ MOOCs for which we trained automatic dropout detectors, along with their academic
field (social science, humanities, STEM, and health sciences), number of participants (registrants
who entered the MOOC courseware at least once), and number of certifiers. The total number of
\emph{unique} participants over all 40 MOOCs was $528349$.
}
\label{tbl:mooc_stats}
\end{table}

\subsection{Target labels}
\label{sec:targets}
The binary target labels used for training and evaluation were whether ($1$) or not ($0$) each student accrued enough points during the MOOC to earn a
certificate. The grade threshold for certification differed across the MOOCs but is typically around $70\%$.
Note that, starting in late 2015, some HarvardX MOOCs implemented a policy whereby only students who paid a fee to have their identity verified could 
officially earn a certificate. For these
courses, we still considered the target label for a student to be $1$ as long as her/his point total exceeded the verification threshold -- in other words,
we ignored the fact of whether or not the student paid money to become ID-verified.

\subsection{Features}
\label{sec:features}
\begin{table}
\begin{tabular}{ll}\hline
\multicolumn{2}{c}{\bf List of person\_course features}\\\hline
LoE & YoB \\
gender & continent \\\hline
\multicolumn{2}{c}{\bf List of person\_course\_day features}\\\hline
avg\_dt &  sdv\_dt \\
max\_dt & n\_dt \\
sum\_dt &  nevents \\
nprogcheck &  nshow\_answer \\
nvideo & nproblem\_check \\
nforum &  ntranscript \\
nseq\_goto &  nseek\_video \\
npause\_video & nvideos\_viewed \\
nvideos\_watched\_sec &  nforum\_reads \\
nforum\_posts & nforum\_threads \\
nproblems\_answered & nproblems\_attempted \\
nproblems\_multiplechoice & nproblems\_choice \\
problems\_numerical & nproblems\_option \\
problems\_custom & nproblems\_string \\
problems\_mixed & nproblems\_formula \\
problems\_other & \\\hline
\multicolumn{2}{c}{\bf List of additional features}\\\hline
precourse\_survey & days\_since\_last\_action\\
\end{tabular}
\caption{Features extracted for MOOC dropout detection for the various training paradigms.
}
\label{tbl:features}
\end{table}

The features that were used for MOOC dropout prediction are 
\emph{clickstream features} that are computed from the clickstream log that contains all interaction events between every student
and the MOOC courseware; this includes answers to quiz questions, play/pause/rewind events on  lecture videos, reading and writing
to the discussion form (the events, not the actual text), and more. The features we chose as the basis for automatic MOOC dropout
detection are similar to prior approaches (e.g., \cite{boyer2015transfer}) and can generalize to a wide variety
of MOOCs across academic disciplines; they  are
listed in Table \ref{tbl:features}.

\subsubsection{Feature extraction}
All features were extracted from two different
database tables  -- \begin{tt}person\_course\end{tt} and \begin{tt}person\_course\_day\end{tt} -- 
that are automatically updated daily by the \begin{tt}edx2bigquery\end{tt} data management framework \cite{ChuangLopez2016} that is used for MITx and HarvardX MOOCs.
The \begin{tt}person\_course\end{tt} features consist of the student's self-reported highest level of education (LoE), year of birth (YoB),
gender, and continent (Africa, Europe, etc.). Each of these variables was converted into a vector of binary dummy variables:
\begin{itemize}
\item Age: dummy variables based on the student’s approximate age (computed as $2012 - \textrm{YoB}$) in the following ranges: $<10$, $10-15$, $15-20$, $20-25$, $25-30$, $30-35$,
$35-40$, $40-45$, $45-50$, $50-55$, $55-60$, $>60$ years, and null (no response).
\item LoE: elementary school, junior high school, high school, associate's degree, bachelor's degree, master's degree, professional degree, and null (no response).
\item Gender: male, female, other, and null (no response).
\item Continent: Europe, Oceania, Africa, Asia, Americas, North America, South America, and null (no response).
\end{itemize}

The \begin{tt}person\_course\_day\end{tt} dataset was used to compute clickstream features 
such as the average and total amount of time spent in the course on a particular day (avg\_dt,
sum\_dt respectively), total number of clickstream events (nevents), etc.  Although \begin{tt}person\_course\_day\end{tt}
contains separate information for each student for each day of the course, we chose to aggregate information across time:
Specifically, when making a dropout prediction  for a specific student
at time $t$, we summed the values within each feature across all days $t' \leq t$. This approach
yields a feature representation that is both smaller and whose dimensionality does not change as a function of  time.\footnote{Note that, in pilot
experimentation we also tried using separate \begin{tt}person\_course\_day\end{tt} features for each day, but the accuracy was lower.}

Finally, we extracted two additional features: \begin{tt}precourse\_survey\end{tt},
which indicates whether or not the student responded to the course's pre-course survey (which might be positively correlated with commitment to the MOOC),
and the number of days since the student interacted with the MOOC courseware (which can be computed from data within \begin{tt}person\_course\_day\end{tt}).
In total (including all dummy variables and both demographic and tracking log features), there were $73$ features used for prediction.

\section{Training Paradigms}
We compared several different paradigms for training automatic MOOC dropout predictors:
\begin{enumerate}
\item \emph{Train on same course (post-hoc)}:
When predicting which students from course $c$ will drop out, train using features and target labels 
from the exact same course $c$. Note that, since target labels for $c$ become available only \emph{after} $c$ has ended, this approach
essentially would require either that (a) the practitioner go ``back in time'' to when the MOOC first started,
or (b) that a new MOOC with the exact same distribution of students (demographics, prior knowledge, etc.) and with the exact
same content and structure is offered in the future, and that no exogenous factors (e.g., comedian Stephen Colbert talking
on television about MOOCs \cite{ho2014harvardx}) cause students to behave differently during the later incarnation of the course. 
These assumptions are unlikely to hold. This approach is, however, a useful benchmark to compare against other training paradigms.
\item \emph{Train on other course from same field}: When predicting which students from course $c$ will drop out, train using features and target labels
from a different course $c'$ that has already completed, and for which the target labels  are thus already available. Although it is difficult to know
\emph{which} prior course should be used for training, a reasonable choice is a different course from within the same discipline (social sciences, humanities,
etc.). We chose to use the \emph{largest} such course in order to maximize the size of the dataset available for training.
\item \emph{Train on many other courses}: When predicting which students from course $c$ will drop out, train using features and target labels
from \emph{many} different courses (not necessarily within the same discipline). Specifically, for each course $c$, we trained a dropout classifier from each of
the $39$ other courses (recall that we tested $40$ MOOCs in total) and then averaged the classifiers' hyperplanes together.
\end{enumerate}

In addition, we also explore a novel training approach, similar to the \emph{in-situ} training method proposed
by \cite{boyer2015transfer}, based on using \emph{proxy labels}:
\begin{enumerate}
\item[4.]
\emph{Train using proxy labels (in-situ)}: When predicting at week $w$ which students from course $c$ will drop out,  train
using \emph{proxy labels} corresponding to whether each student \emph{persisted} -- i.e., interacted with the MOOC courseware at least once -- within the previous week $w-1$ (see Figure \ref{fig:insitu}).
\begin{figure}
\begin{center}
\includegraphics[width=3.3in]{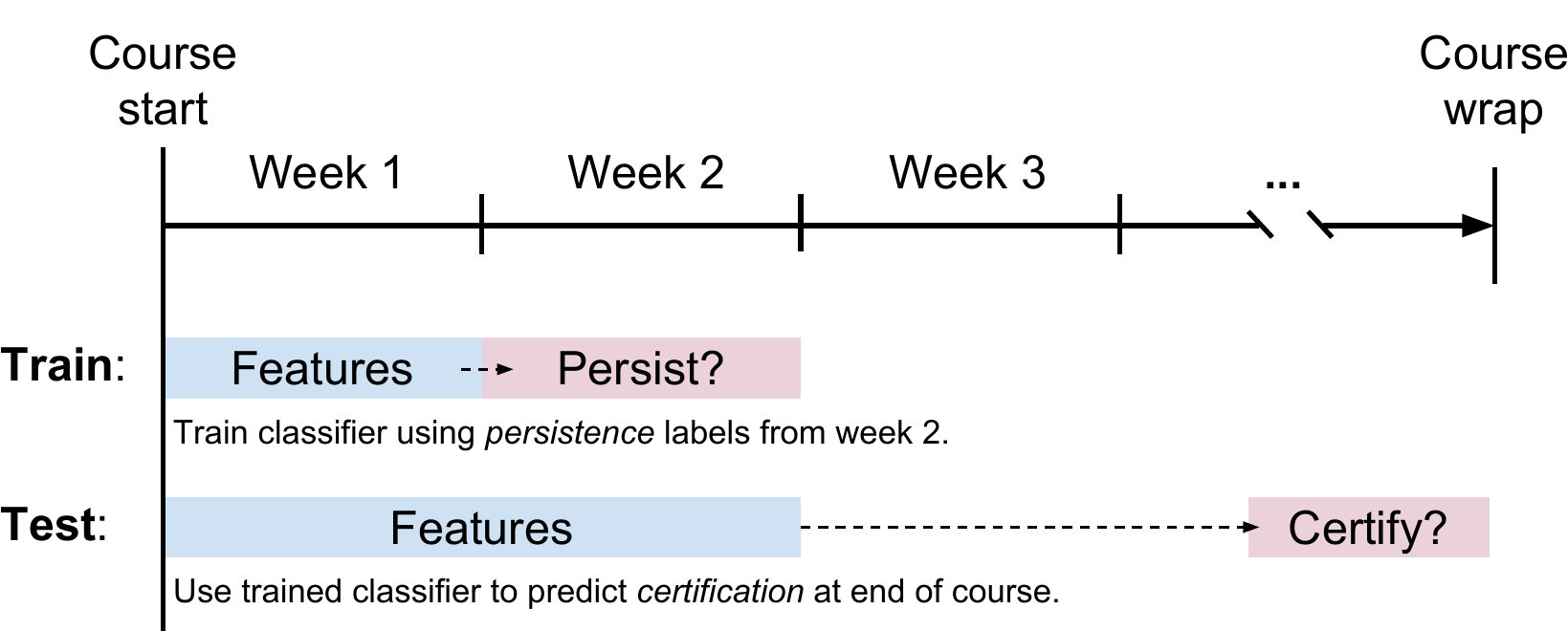}
\end{center}
\caption{Schematic representing how to Train using proxy labels (in-situ): At each week $w$,
proxy labels of whether the persistent \emph{persisted} during week $w-1$ are used to predict
whether or not the student will certify/drop out by the end of the course.}
\label{fig:insitu}
\end{figure}
This approach can be implemented for any MOOC, and because it  does not require ``seeing into the future'' to obtain target labels,
it can deliver a dropout predictor that can be deployed during a \emph{live} MOOC, not just after it has finished.
There is, however, an inherent
mismatch between the \emph{training} target labels (based on persistence)
and the \emph{testing} target labels (based on certification/dropout), and this mismatch may degrade performance.
\end{enumerate}

\section{Classification architecture}
The classification architecture for the detectors was  logistic regression with $L_2$ regularization, which is equivalent to a 2-layer
neural network.  
When estimating parameters of the logistic regression models, we varied the strength of the regularization parameter over the set $C \in \{ 10^{-4}, 10^{-3}, 10^{-2}, 10^{-1}, 10^0, 10^1, 10^2 \}$. Prediction accuracy on the test set did not substantially change (relative to the package-default value of $C=1$) over this set, and we thus report all experimental results for $C=1$.
We used the open-source \begin{tt}sklearn\end{tt} and \begin{tt}numpy\end{tt} packages for training.

\subsection{Baseline approaches}
In order to gauge how much ``added value'' is brought by machine learning approaches to MOOC dropout prediction that utilize
detailed clickstream information, we also compared the architectures described above to two simple baseline heuristics.
In particular, \emph{Baseline 1} uses only demographic information -- consisting of self-reported year of birth, continent of origin (Africa, North America, etc.),
level of education (primary/elementary school, high/secondary school, college, etc.), and gender -- to make predictions. These information
are available for each student as soon as she/he registers for the course; hence, no clickstream data is required. As with the other approaches,
logistic ridge regression was used for classifier training; we trained a separate logistic regression classifier for each course.

We also compared against an even simpler \emph{Baseline 2} which requires no machine learning
at all; rather, the predictor makes predictions based on the number of days since the student last interacted with the courseware.
In prior research on dropout detection \cite{WhitehillEtAl2015,KizilcecHalawa2015}, this variable alone has shown to be highly predictive of dropout.

\subsection{Normalization}
For {\em Train on same course (post-hoc)}, {\em Train on other course from same field}, and {\em Train on many other courses},
both the training and testing sets were normalized by subtracting each feature value by the feature-wise mean and dividing by the
feature-wise standard deviation, where the mean and standard deviation were calculated over only the training data. This normalization
is important to ensure that the L2 regularization affects all features similarly. 
For {\em Train using proxy labels (in-situ)}, we used a different type of normalization to ensure that feature values are comparable
even when they are extracted across time windows of different lengths (see Figure 1). Specifically, we converted all clickstream
features from absolute counts to percentiles.

\subsection{Accuracy metric}
To measure accuracy of the dropout classifiers, we use the Area Under the receiver operating characteristics Curve (AUC) metric. The Receiver Operating Characteristics (ROC) curve itself plots the true positive rate versus the false positive rate of the trained classifier. The AUC is the integral of the ROC curve over the interval $[0,1]$ and answers the following question: Given one randomly chosen student who drops out from the MOOC and one randomly chosen student who certifies in the MOOC, what is the probability that the classifier can correctly distinguish the two students? A useless classifier that simply flips a coin will have an AUC of 0.5, whereas a perfect classifier will have an AUC of 1. (A classifier with AUC of 0 always makes the wrong discrimination.) Importantly, dropout accuracy as assessed by the AUC statistic is not affected by the overall proportion of students within a particular course who drop out (which is typically very high in MOOCs).

\subsection{When to Measure Accuracy}
In order to obtain reliable estimates of the accuracy of a MOOC dropout detector, it is important to consider over what time period the
accuracy is computed. For the goal of automated intervention, it is more useful to be able to predict early in the course, rather than later, which students
will eventually drop out.  Another point to consider is that, near the end of the MOOC, some students may have already accrued enough points
to earn a certificate. A dropout detector that predicts that such students will not drop out is not so much ``predicting'' these students'
future performance as it is reporting what they have already achieved; hence, accuracy statistics computed over such students may overestimate
the performance of the predictor. For both these reasons, we decided to compute accuracy over all weeks of each MOOC between the course launch
date ($T_{0\%}$, when instruction begins) and the earliest date by which students could possibly have earned enough points to earn a certificate ($T_{100\%}$).

Note that the earliest time at which predictions can be made differs across the training paradigms. Specifically, the \emph{Train on same course (post-hoc)}
approach can make predictions after just the first week of the course. In contrast, \emph{Train using proxy labels (in-situ)} 
needs at least 2 weeks of clickstream data before the first set of proxy labels become available.  Simple baselines based on students' demographics that
require no clickstream data can be applied as soon as students register, which is a notable advantage compared to the machine learning-based
approaches.

\section{Results}
\begin{figure}
\begin{center}
\includegraphics[width=3.35in]{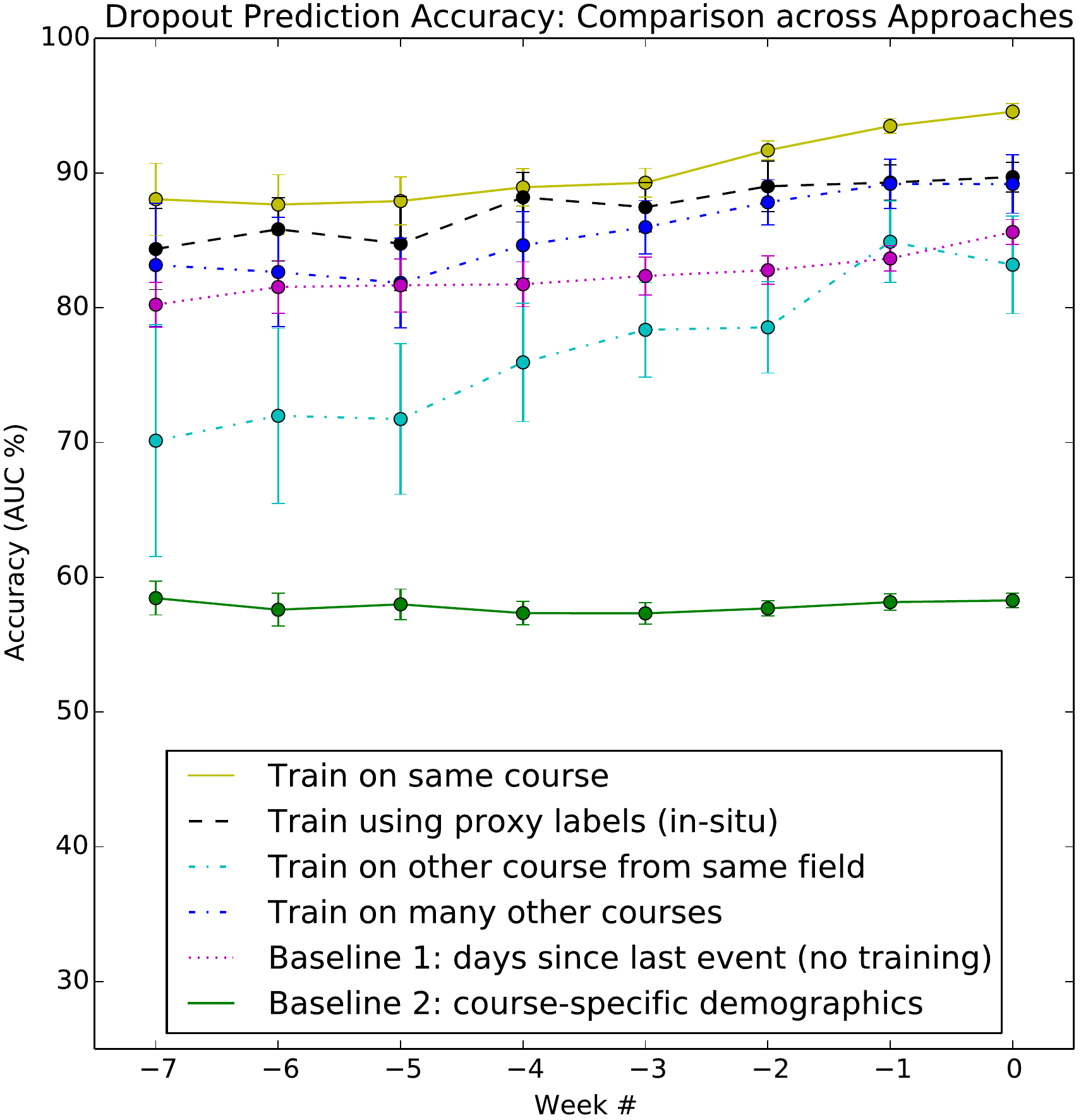}
\end{center}
\caption{Results comparing different approaches.}
\label{fig:results}
\end{figure}
Dropout prediction accuracy (AUC) for all training approaches are shown in Figure
\ref{fig:results}. The horizontal axis indexes the week in the MOOC, where week 0 is defined for each course to be $T_{100\%}$. (Week $-3$
corresponds to 3 weeks prior to $T_{100\%}$, etc.) Each data point is shown with error bars corresponding to the standard error
of the mean. Since the lengths of the courses varied, the accuracy statistics for each week $w$ were computed using only those
MOOCs for which data were available at week $w$.

As shown in the graph, the most accurate prediction paradigm was \emph{Train on same course (post-hoc)},
which is the predominant training paradigm used in most prior dropout prediction research.  It
achieved an accuracy (averaged over all 8 weeks, and all MOOCs within each week) of $90.20\%$.
Perhaps more surprising is that the second most accurate approach was \emph{Train using proxy labels (in-situ)}. This approach
does not require any MOOC -- similar or dissimilar -- to have been offered previously. Despite the inherent mismatch between
the labels of persistence (did the student participate during the previous week?) and labels of certification (will the student
earn enough points to earn a certificate?), this approach attained an accuracy (averaged over all tested MOOCs and all 8 weeks) of $87.33\%$.

The third most accurate approach was \emph{Train on many other courses}, with an accuracy of $85.56\%$.
This training paradigm attained a higher accuracy than did \emph{Train on other course from same field} ($76.85\%$), suggesting that,
if it is not possible to exploit course-specific structure via either \emph{Train on same course (post-hoc)} or \emph{Train using proxy labels (in-situ)},
then it is better to harness prior data from a large variety of courses than from just a single course (even from within the same discipline).

\section{Comparison to Baseline Heuristics}
\emph{Baseline 1}, whose predictions are based solely on each student's self-reported demographics,
achieved an average prediction accuracy of $58.85\%$, suggesting that only a small amount of information about dropout is contained
in the demographics. (Note that we also tried a demographics-based classifier whose parameters were averaged over many courses, rather
than just for the one course itself; this approach's accuracy was even lower.) 

\emph{Baseline 2}, whose predictions are solely on the number of days since the student's last interaction with the course, performed remarkably well:
It attained an average dropout prediction accuracy of $82.45\%$, which corroborates previous findings \cite{WhitehillEtAl2015,KizilcecHalawa2015}
that this variable is highly salient for prediction. Nonetheless, \emph{Baseline 2} was still substantially less accurate than either
\emph{Train on same course (post-hoc)} or \emph{Train using proxy labels (in-situ)}, suggesting that harnessing more detailed
clickstream features does bring a substantial accuracy boost.



\section{Accuracy across academic fields}
To explore whether dropout prediction accuracy tended to be higher for MOOCs within a particular academic field -- STEM, Humanities, Social
Sciences, or Health Sciences -- we implemented a linear mixed-effects model in which dropout prediction accuracy for the 
\emph{Train on same course (post-hoc)} paradigm was estimated using
the week $w$ and field $f$ as fixed effects and the course
as a random effect.

{\bf Results}: The week $w$ was statistically significantly ($\chi^2(1)=184.77$, $p<0.0001$) correlated
with dropout prediction accuracy: for each week later in the course (i.e., closer to $T_{100\%}$), there was an estimated $1.53\%$ increase in prediction accuracy.
The difference in average accuracy varied only slightly as a function of the field: the prediction accuracy tended to be
slightly lower for STEM courses and slightly higher for humanities courses. However, the difference between the intercepts for these fields
was only $2.19\% - (-.16\%) = 2.35\%$; it was not statistically significant ($\chi^2(3)=1.9825$, $p=0.576$). This suggests that the features
listed in Table \ref{tbl:features} that we extract for dropout prediction are robust to changes in the MOOC subject matter.

\section{Deeper Prediction Architectures}
Most of the prior research on automatic MOOC dropout prediction has used generalized linear models -- including
logistic regression, support vector machines, and survival/hazard models -- to classify clickstream, natural language, and/or social
network features (see Table \ref{tbl:lit_review}). Only a handful \cite{fei2015temporal,ColemanEtAl2015,BalakrishnanBerkeley2013,xing2016temporal}
of approaches have considered
deeper learning architectures that can capitalize on nonlinear feature representations and interactions between features.
In this section, we explore the potential benefits of employing a deep, fully-connected, feed-forward neural network
for dropout prediction. Specifically, we performed an experiment in which we systematically vary both the depth
(number of hidden layers), as well as the width (number of neurons per hidden layer), of a neural network,
whose $73$ input features are the same as in Section \ref{sec:features}. Between the input and the first hidden layer,
and between each pair of consecutive hidden layers, are rectified linear units (ReLUs). The final, output layer
is a softmax layer.
The course we used for experimentation was the  MOOC with the greatest number of certifiers -- GSE2X (see Table \ref{tbl:mooc_stats}),
and we trained neural networks to predict dropout during the last week of the course prior to $T_{100\%}$.

To make experimentation more efficient, we utilized
a recently developed methodology for neural network training called Net2Net \cite{Net2Net2016}.
Net2Net is useful when training a sequence of  incrementally more complex networks.
In particular, with Net2Net there exists a \emph{teacher} network from which a new \emph{student} network
is built. The student network is deeper or wider than the teacher network, but the weights of the student network are initialized
prior to training either by (a) replicating neurons within a hidden layer and adjusting the weights to/from those neurons
to compensate (Net2Wider); or (b) initializing the weights of the neurons in an additional hidden layer to implement the identity function
(Net2Deeper). In this manner, the student network is guaranteed to produce identical outputs as the teacher network at the start
of the training and can become even more accurate during training.

\begin{figure}
\begin{center}
\includegraphics[width=3.35in]{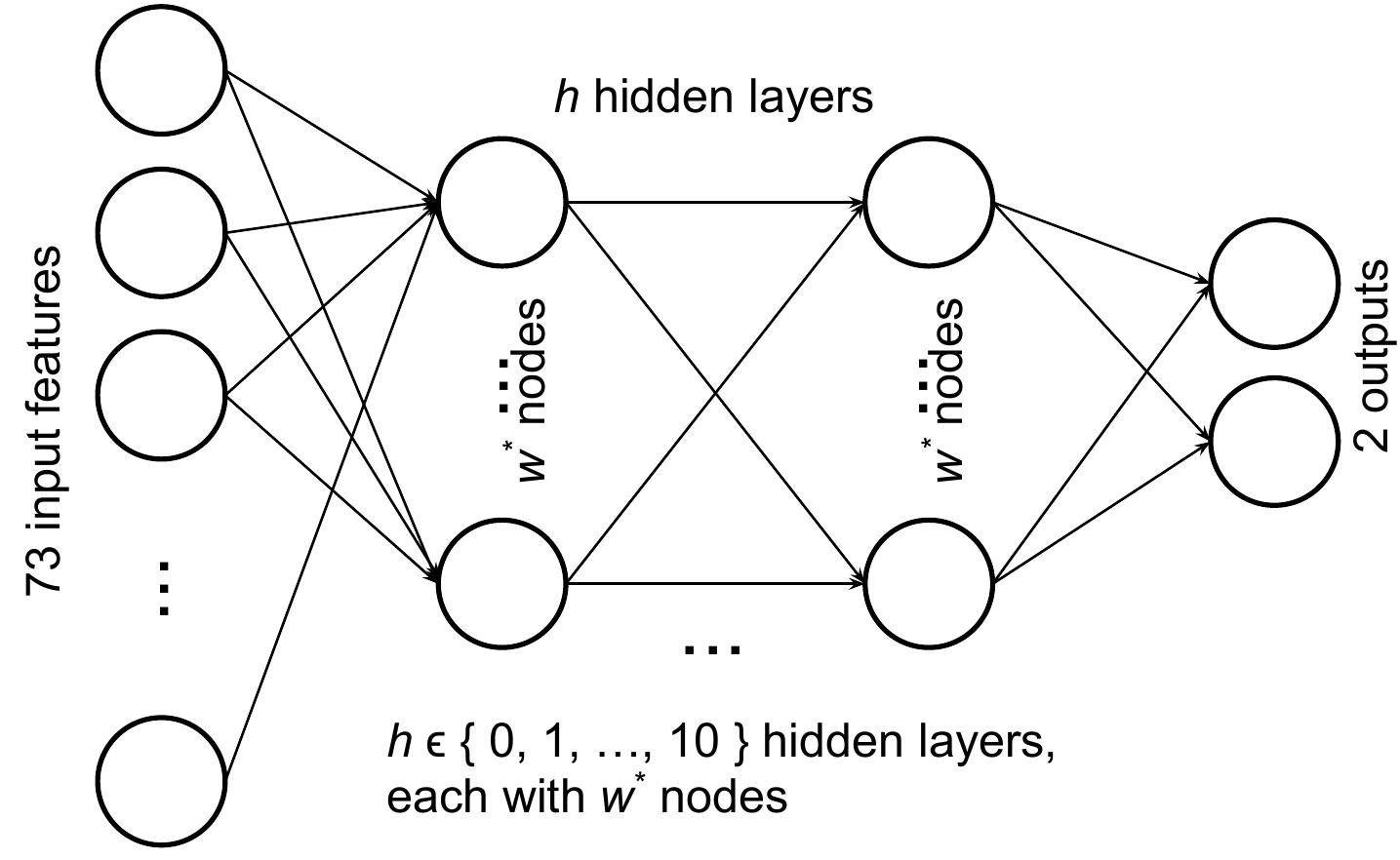}
\end{center}
\caption{Architecture of fully-connected feed-forward neural network for MOOC dropout detection. Each of the $h$ hidden layers 
has $w^*=5$ neurons.}
\label{fig:net}
\end{figure}

Using Net2Net, we varied the number of hidden layers $h$ as well as the number of neurons per hidden layer $w$ -- see 
Figure \ref{fig:net}. We began with $h=1,w=2$, i.e., a 3-layer neural network with 2 neurons in the hidden layer.
We then applied the Net2Wider technique using this network as the teacher and built a student network with
$w=3$ neurons  in the hidden layer. Next, the $w=3$ neuron network was used as the teacher,
and a network with $w=4$ neurons was constructed. This procedure was carried out iteratively until $w=15$ was reached,
such that a $w$-neuron student was trained from an $w-1$-neuron teacher, where $w \in \{3,4,\ldots,15\}$. The prediction accuracy
at each value of  $w$ was measured on test data; while the highest accuracy was attained for $w=6$, accuracy at 
$w=5$ was very similar (and requires fewer parameters).
After maximizing accuracy on test data for $h=1$ with respect to $w$, we fixed $w^*=5$ and then applied the Net2Deeper approach
to maximize accuracy with respect to $h$. Specifically, we iteratively trained a student network with $h$ hidden layers from
a teacher hetwork with $h-1$ hidden layers, where $h\in\{2,3,\ldots,10\}$.

Each network was trained using stochastic gradient descent with no momentum, a learning rate of $0.1$,
over 20 epochs. The minibatch size was 10, and  the learning rate was annealed by a factor of $1/(1+10^{-3})$
after training on each mini-batch. To account for class imbalance, we tried
re-weighting the loss of the two classes; however, this technique did not increase test accuracy and was thus dropped. We used
the \begin{tt}keras\end{tt} software package for training.

\subsection{Results}
\begin{figure}
\begin{center}
\includegraphics[width=3.35in]{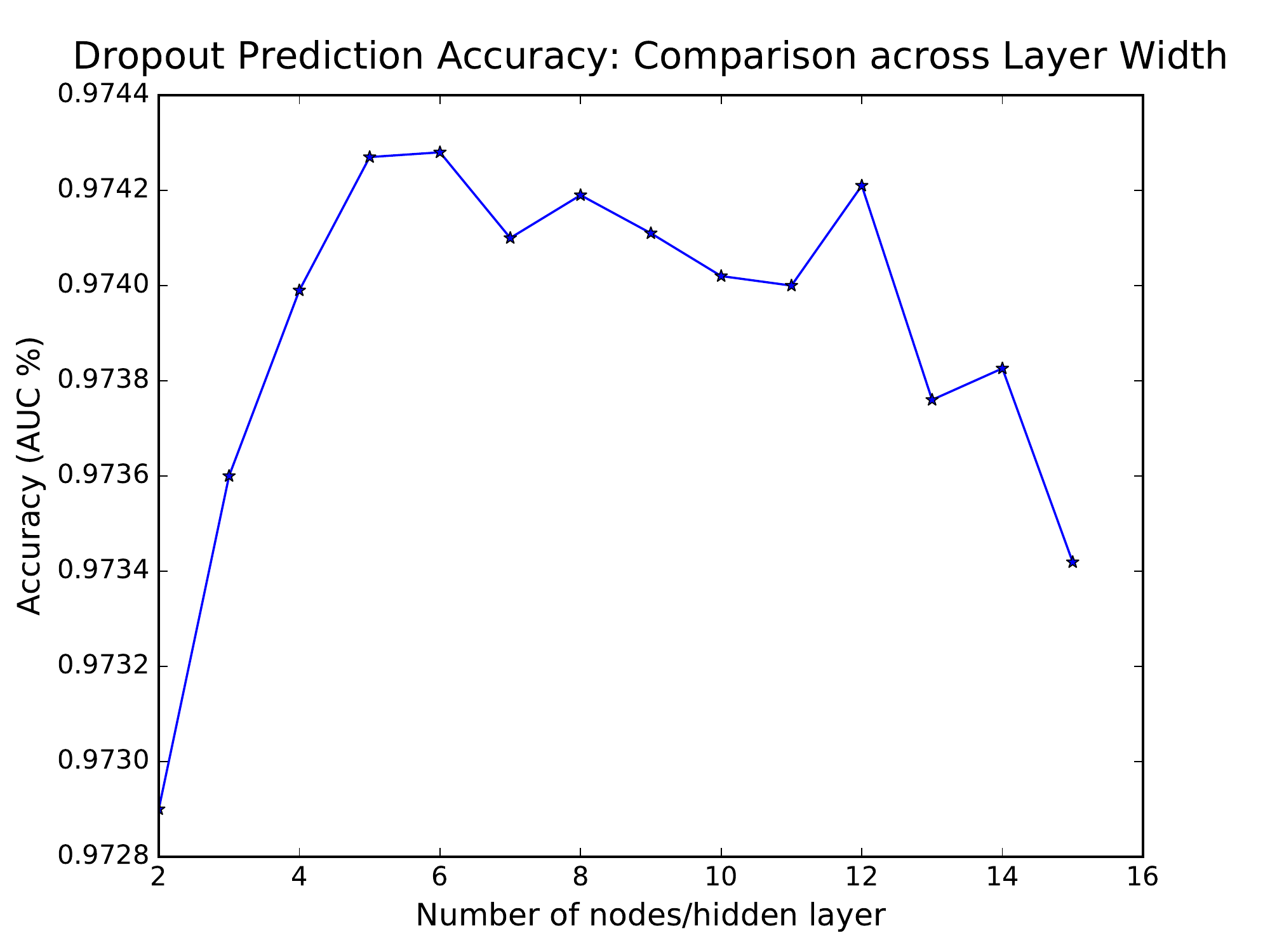}\\
\includegraphics[width=3.35in]{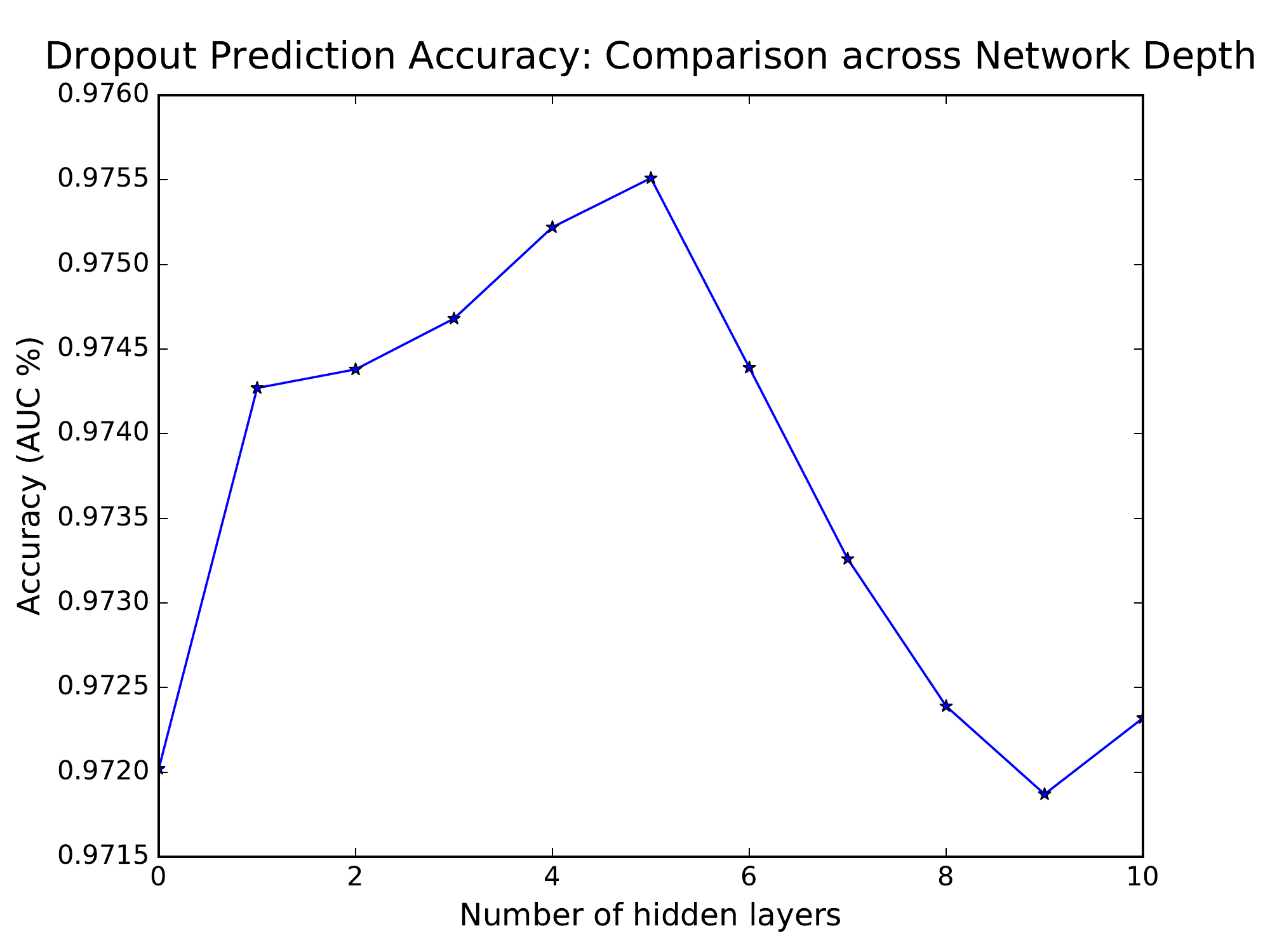}
\end{center}
\caption{{\bf Top}: MOOC dropout prediction accuracy, on test data from GSE2x, of a neural network (with one hidden layer) as a function of the number of nodes in
the hidden layer. {\bf Bottom}: Accuracy of a neural network (with 5 nodes per hidden layer) as a function of the number of hidden layers.}
\label{fig:deep_net_results}
\end{figure}

Results of maximizing both respect to $w$ and (separately, after fixing $w^*=5$) $h$ are shown in Figure \ref{fig:deep_net_results}.
For a network with $h=1$ hidden layer, the accuracy at $w=5$ was $97.43\%$. When keeping the width fixed at $w^*$,
the highest accuracy -- $97.55\%$ -- was achieved for $h=5$. For comparison,  logistic regression -- which is equivalent to
a neural network with $h=0$ hidden layers --  achieved an accuracy of $97.20\%$. The difference between these accuracies -- $97.55 - 97.20 = 0.35\%$ --
was statistically significant ($t(23738) = 78.47, p<0.00001$, two-tailed). For comparison, the difference in accuracies between the first-place
and second-place contestants in the 2015 KDD Cup contest on MOOC dropout prediction was $90.92 - 90.89= 0.03\%$ \cite{KDDCup2015}.

{\bf Training cost}:
On GSE2x, whose training set contains $11870$ examples, and using the Net2Net training technique, 
it took 80 seconds (an average of 4 seconds per epoch) for the best network ($w=5,h=5$) to train on an NVIDIA Tesla K20Xm GPU.

\section{Summary and conclusions}
In this paper we have investigated the \emph{practical} level of accuracy that can be achieved with an automatic MOOC
dropout predictor, given the fact that detectors must be trained before they can be deployed, using training data that are
collected \emph{before} their first deployment. In particular, we have compared several different training paradigms that are widely
used in the MOOC dropout prediction literature -- e.g., train on the same course (\emph{post-hoc}), train on different course
from same academic discipline, and train on many different courses -- on 40 different MOOCs that span a variety of
disciplines including the humanities, social sciences, health sciences, and STEM. Results suggest that the accuracy of classifiers
that are trained on data that are collected only \emph{after} a course has finished -- and which are thus unusable in the MOOC itself --
are often several percentage points higher than classifiers that are trained on other MOOCs. This underlines the importance of
careful accuracy estimation before conducting a large-scale intervention.

In addition, we have explored a training approach, similar to \emph{in-situ} training \cite{boyer2015transfer},
based on the idea of \emph{proxy labels} -- labels that approximate the quantity of interest (dropout versus certification)
but that can be collected \emph{before} a course has completed. Surprisingly, the accuracy of this approach is very similar to when
classifier are trained \emph{post-hoc} from data of courses that have already finished.
Moreover, this approach  enables a MOOC dropout classifier both to be trained and to be deployed while a course is ongoing. 

Finally, we have presented evidence that deeper feature representations of basic demographic and clickstream features -- as implemented
with a deep (7 layers including 5 hidden layers with 5 neurons per hidden layer), fully-connected, feed-forward neural network can statistically
significantly increase the accuracy of MOOC dropout prediction relative to generalized linear models that
are the standard today. Using an iterative training strategy \cite{Net2Net2016},
the total training time to create such a network is modest and amenable to large-scale implementation over many (hundreds) of MOOCs
at regular training intervals (e.g., once per week).

{\bf Future work} on dropout detection should focus on moving from prediction to intervention. In particular,
more research should be invested in identifying not only those students who will drop out, but specifically those
who are likely to benefit most from interventions.

\bibliographystyle{abbrv}
\bibliography{paper}  

\end{document}